\documentclass[%
 sor,
 jor,
 amsmath,amssymb,
 reprint,%
]{revtex4-2}

\usepackage{graphicx}
\usepackage{multirow}
\usepackage{caption}
\usepackage{subcaption}
\usepackage{dcolumn}
\usepackage{bm}
\begin{document}
\renewcommand{\abstractname}{Abstract}

\preprint{AIP/123-QED}

\title{Deep Learning to advance the Eigenspace Perturbation Method for Turbulence Model Uncertainty Quantification}

\author{Khashayar Nobarani}

\affiliation{ 
Department of Mechanical Engineering, University of Tabriz, Iran
}%

\author{ Seyed Esmaeil Razavi}

\affiliation{ 
Department of Mechanical Engineering, University of Tabriz, Iran
}%

\date{\today}

\begin{abstract}
The Reynolds Averaged Navier Stokes (RANS) models are the most common form of model in turbulence simulations. They are used to calculate Reynolds stress tensor and give robust results for engineering flows. But RANS model predictions have large error and uncertainty. In past, there has been some work towards using data-driven methods to increase their accuracy. In this work we outline a machine learning approach to aid the use of the Eigenspace Perturbation Method to predict the uncertainty in the turbulence model prediction. We use a trained neural network to predict the discrepancy in the shape of the RANS predicted Reynolds stress ellipsoid. We apply the model to a number of turbulent flows and demonstrate how the approach correctly identifies the regions in which modeling errors occur when compared to direct numerical simulation (DNS), large eddy simulation (LES) or experimental results from previous works.
\end{abstract}

\maketitle

\section{\label{sec:level1}Introduction}

In engineering design and analysis we deal with problems and applications that involve turbulent flow. Engineers and scientists are always looking for methods to simulate and predict turbulent flows so that they get good accuracy on the results, and at the same time the methods are not computationally expensive. One of the most accurate strategies is direct numerical simulations (DNS), but solving engineering problem with DNS is time consuming and computationally expensive. So researchers need some alternative ways to deal with turbulent flow problems. Because of the complexity of these engineering problems we have to simulate them using turbulence model based computational fluid dynamics (CFD) simulations. 

The most popular approach which has been developed by researchers are Reynolds Averaged Navier Stokes (RANS) models. In RANS simulation instead of fully resolving turbulent motions, we aim to find the average velocity and pressure fields and to model the effects of fluctuating components. The averaging means that we only resolve the large scales of motions and model the averaged effect of the smaller turbulent scales of motions. This is differennt from DNS where all scales of motion are resolved and LES where only the smallest scales of motion are unresolved. The effects of fluctuating turbulent scales arise in a term in Reynolds Averaged Navier Stokes equation which is called Reynolds stress. The goal of RANS models are to determine Reynolds stress in term of average component. Although more traditional two equation models are commonly used in engineering problem such as $k-\epsilon$ and $k-\omega$ models, their suitability to flows with different features are limited \citep{speziale1991analytical,johansson2002engineering,chen2003extended}. RANS models are less computationally expensive, but in flows with features such as streamline curvature, separation and high adverse pressure gradient, RANS models are very inaccurate. These features are very common in complex flows found in engineering. For example the diffusers used inn turbomachinery have turbulent flow and show turbulent flow separation in the expansion sections along the walls of the diffuser. Because of the inaccuracies of RANS models it is required to determine the errors and uncertainties in their predictions. Uncertainty quantification for RANS models attempts to assess the potential errors in model prediction and is important in establishing RANS model's reliability in engineering problems.

In the past few years the advancement of data science and  machine learning algorithms in daily life task and engineering problems such as pattern recognition (\cite{10.1117/1.2819119}), language processing (\cite{collobert2008unified}), and voice detection have had large impact across fields (\cite{waller2013data}) and have been impactful in the tech industry.
Recently, scientists and engineers have started to explore many different techniques from Machine Learning for scientific tasks such as DNA sequencing (\cite{libbrecht2015machine}) and the discovery of new materials (\cite{pilania2013accelerating}), astronomy (\cite{baron2019machine}).

Fluid mechanics researchers annd turbulence modellers have also started to develop Machine Learning techniques for fluid mechanics. One of the first problem to be tackled by machine learning is to inspect RANS inaccuracies and later on to come up with a more sophisticated model for Reynolds stresses. \cite{singh2016using} used a data-driven method in which  data sets are used to conclude the functional form of model discrepancies. This was augmented by machine learning algorithms to reconstruct model corrections.
\cite{ling2015evaluation} developed a marker using random forest, Adaboost decision tree and support vector machine to determine the regions where RANS simulations are inaccurate without making quantitative estimate of uncertainty in RANS prediction. \cite{ling2016reynolds} proposed the Tenor Basis Neural Network  to learn the coefficients of an integrity basis for the Reynolds anisotropy tensor which was suggested by \cite{pope1975more}. (\cite{fang2018deep} who was inspired by \cite{ling2016reynolds} work, used machine learning to model the Reynolds stress for fully-developed turbulent channel flow. They used a fully-connected neural network as the base model and increased its capabilities by using non-local features, directly implementing friction Reynolds number information into the machine learning process. Machine Learning based applications in turbulence modeling have both advantages and disadvantages. An advantage is that Machine Learning based models are not encumbered by the limitations of the human expert and can derive information from very large amounts of high-dimensional data. A major disadvantage is the non-generalizability of Machine Learning based models. Machine Learning based turbulence models are accurate only for flows that a similar to the flows they have been trained upon. This non-generalizability can be reduced by adding physics based knowledge into the Machine Learning model formulation as physics based laws are universal. A combination of a physics based framework with Machine Learning based algorithm may give the best results that are accurate and still generalize beyond the flows used to train the model.

The only physics based framework for uncertainty estimation is the Eigenspace Perturbation Method developed by \cite{iaccarino2017eigenspace}. This method uses physics based perturbations of the RANS model's predicted Reynolds stress eigenvalues, eigenvectors and turbulent kinetic energy. Each kind of perturbation extends the isotropic eddy viscosity based model to a general anisotropic eddy viscosity model \citep{mishra2019theoretical}. The perturbations lead to a set of different predictions of turbulence quantities. The union of these predictions gives a interval estimate for the model form uncertainty from the RANS model. The Eigenspace Perturbation Method has been applied to engineering design problems with success \citep{ cook2019optimization,granados2019influence,mishra2020design,razaaly2019optimization, mishra2017uncertainty}. But there are some limitations in the Eigenspace Perturbation Method. The Eigenspace Perturbation Method is based on physics based constraints and bounds to determine Reynolds stress perturbations for uncertainty estimation. These physics based constraints only inform about physical possibility and not about physical plausibility. Physics based bounds tell us how to perturb the Reynolds stress tensor and the range of physically allowable perturbation. They do not tell us how much to perturb the Reynolds stress tensor for different turbulent flows. This leads to uncertainty bounds that are not as sharp as possible for some turbulent flows\citep{mishra2019uncertainty}.

The magnitude of perturbation for the Eigenspace Perturbation Method can be learned using a Machine Learning model. \cite{heyse2021estimating} trained a random forest model to predict the magnitude of perturbations to the Reynolds stress eigenvalues. In this paper we extend and improve on the work of \cite{heyse2021estimating}. We are using RANS and DNS data, with a neural network to determine how much to perturb at every point in the flow domain for different flows. The deep learning model used in the present paper may lead to better results than the random forest model as the neural networks are universal approximators. The present work focuses determining the amount of perturbation needed for uncertainty quantification. We focus on a data-driven framework based on an extension of the eigenvalue perturbation to the Reynolds stress anisotropic tensor developed. The complete analysis of the discrepancy needs perturbations to the eigenvalues, the eigenvectors and the turbulent kinetic energy. In this paper we keep our analysis and modeling to the eigenvalue perturbation. Our data-driven framework is used to predict the magnitude of eigenvalue perturbation is needed at every point in the flow domain. The neural network model is used to learn a mapping from the input features to the eigenvalue perturbation magnnitude. These predictions are compared with the DNS and experimental data from previous works.

\section{\label{sec:level2}Computational Details}
\subsection{\label{subsec:level1}Details of RANS Modeling}

The Reynolds-average Navier-stokes approach splits the velocity and pressure fields into a mean and a fluctuating component:
\begin{eqnarray}
\label{eq:1}
u = \overline{u} + u^\prime\\
\label{eq:2}
p = \overline{p} + p^\prime
\end{eqnarray}
Implementing equation \ref{eq:1} and \ref{eq:2} into the Navier Stokes equations we get:
\begin{equation}
    \frac{\partial\overline{u}}{\partial t} + \nabla.(\overline{u} \times\overline{u}) = -\frac{1}{\rho}\nabla\overline{p} + \nu\nabla^{2}\overline{u} - \nabla.R
\end{equation}
\begin{equation}
    \nabla.\overline{u} = 0
\end{equation}
where :
\begin{equation}
    R = \overline{u \times u}
\end{equation}
is called Reynolds stress tensor. The Reynolds stress tensor must be modeled to close the RANS equation. A lot of modeling efforts have focused on anisotropy Reynolds stress tensor which is described as:
\begin{equation}
    a = \overline{u \times u} - \frac{2}{3}kI
\end{equation}
As that is the portion responsible for turbulent transport, where k is the turbulent kinetic energy:
\begin{equation}
    k = \frac{1}{2}\overline{u^{\prime} \cdot u^\prime} = \frac{1}{2} trace(\overline{u \times u})
\end{equation}
In this research, the models are trained on the normalized anisotropy tensor, $\textbf b = \frac{a}{2k}$.
Eddy viscosity models such as $k-\epsilon$ and $k-\omega$ are the some of the most used models in engineering problems because of their easy implementation and low cost of computation, but they do not provide satisfactory predictions for all flows \citep{speziale1991analytical,johansson2002engineering,chen2003extended}.

Recently, the machine learning algorithms and especially deep learning have been applied to this problem and obtain prediction for anisotropy Reynolds stress tensor. 

\subsection{\label{sec:level2}Eigenspace Perturbation Method Details}
 \cite{iaccarino2017eigenspace} discovered the Eigenspace Perturbation Method for uncertainty quantification of turbulence models. This method introduced perturbation to eigenvalues, eigenvectors and turbulent kinetic energy of the normalized an isotropy tensor toward limiting states of the turbulence anisotropy. At first we have to decompose the Reynolds stress anisotropy tensor in the form of its eigenvalue and eigenvectors as:
\begin{equation}{\label{eq:8}}
     a_{ij} = v_{im}\Delta_{mn}v_{jn}
\end{equation}
$v_{im}$ is the matrix of orthonormal eigenvector, and $\Delta_{mn}$  is a diagonal tensor with the eigenvalue of $a_{ij}$ on its diagonal.The coordinate system made of the eigenvectors is termed the principal coordinate system in which $a_{ij}$ is diagonal.The $\Delta_{mn}$  matrix have three eigenvalue of $\lambda1,\lambda2,\lambda3$  where, $\lambda1>\lambda2>\lambda3$  and $\lambda1+\lambda2+\lambda3=1$.

\begin{figure}
\includegraphics[scale=1.3]{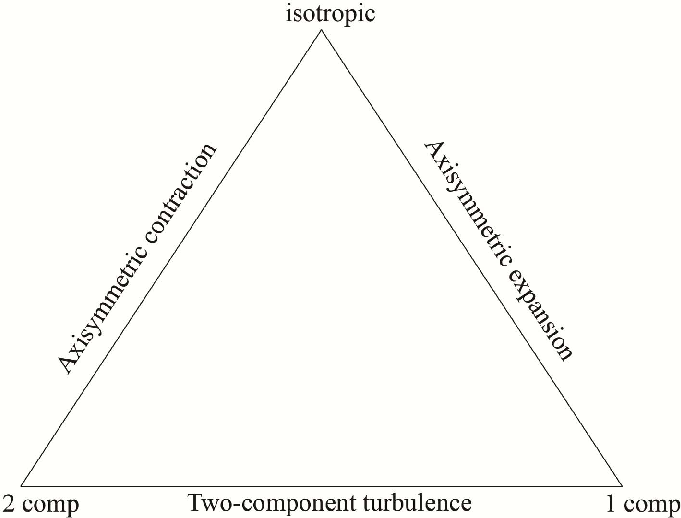}
\caption{\label{fig:Barycentric map} Barycentric map with three limiting state of turbulence anisotropy.}
\end{figure}

These eigenvalues can be visualized as a location in the barycentric map. In that map all realizable state of the turbulent flows are limited within a triangle and the corners of triangle correspond to the limiting state of turbulence with 1, 2 and 3 components respectively as shown in Fig.\ref{fig:Barycentric map}%

Inserting equation (\ref{eq:8}) into Reynolds anisotropic tensor we get:
\begin{equation}
    \textbf b_{ij} = 2k(v_{im}\Delta_{mn}v_{jn} + \frac{\delta_{ij}}{3})
\end{equation}

To account for the errors due to modeling assumptions, the tensor perturbation approach introduced perturbation into the modeled Reynolds stress. This perturbed for expressed as: 
\begin{equation}
    b^*_{ij} = 2k^*(v^*_{im}\Delta^*_{mn}v^*_{jn} + \frac{\delta_{ij}}{3})
\end{equation}
where $*$ represent the perturbed quantities. Thus $k^* = k + \Delta k$ is the perturbed turbulent kinetic energy, $\nu^*_{im}$ is the perturbed eigenvector matrix and $\Delta^*_{mn}$ is the diagonal matrix of perturbed eigenvalues, $\lambda^*_{l}$. In this paper we focus on the eigenvalue perturbations only. Techniques for eigenvector perturbation in the EPM have been applied in earlier research \citep{iaccarino2017eigenspace,thompson2019eigenvector}.

\subsection{\label{subsec:level3}Neural Networks Details}

\begin{figure}

\includegraphics[scale=.7]{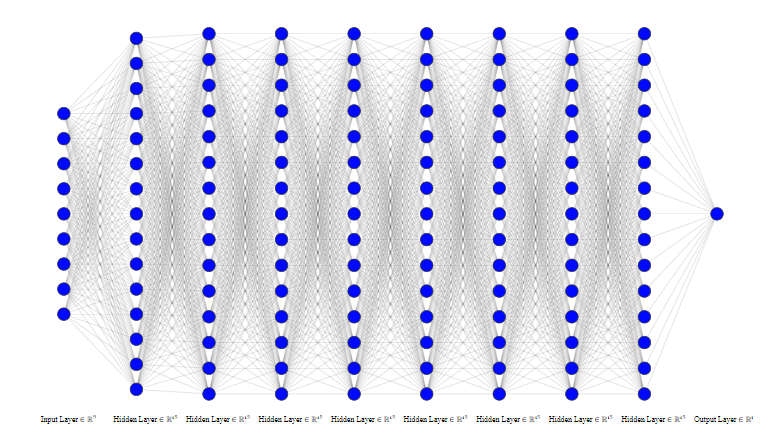}
\caption{\label{fig:Neural Network} Outline of a fully connected neural network}
\end{figure}

Neural networks are a class of machine learning models that have been applied to a lot of problems in variety of fields such as computer vision (\cite{krizhevsky2012imagenet}, natural language processing (\cite{lecun2015deep} and gaming (\cite{silver2017mastering}). Neural networks can easily deal with high-dimensional data and modeling nonlinear and complex relationships. Feedforward neural networks also called multilayer perceptron (MLP) or densely connected neural networks, are one of the primary architectures of deep neural network algorithms. Feedforward neural networks are used to obtain an approximationn of some function  $f$ . Mathematically a neural network defines a mapping from inputs to the targets. The function  $f$  can be a very complex function which will be calculated by the neural network. As an example, Figure \ref{fig:Neural Network} depicts a basic fully connected neural network that defines a mapping from 
The input layer consists of several input vector \{$x_1,x_2 ,x_3,\ldots, x_n $\}. Each component of input vector is called feature which represents an individual property of the flow. The hidden layers perform a nonlinear operation to map the input features to output.

The unknown weight matrix in each layer is estimated using backpropagation approach, as described below. The network is first initialized with constant zero weight for each layer and the error in the network prediction are obtain by comparing them with training datasets or true value y using a user specified analytic cost function such as $L_2$ norm:

\begin{equation}
    E = \sum_{i=1}^{\infty} (y_i - y_t)^{2}
\end{equation}

For the above define analytic cost function, the derivatives of the errors in the output layer with respect to the output feature is:
\begin{equation}
    \frac{\partial E}{\partial y} = 2(y_i - y_t)
\end{equation}
and the derivatives of the errors with respect to the input features is:
\begin{equation}
    \frac{\partial E}{\partial z_i} = \frac{\partial E}{\partial y_i}\frac{\partial y_i}{\partial z_i} = f^{'}(z_{i})\frac{\partial E}{\partial y_{i}}
\end{equation}
where, $f$ is a known analytical function which will be described later on. Since the input features of a layer are linearly related to the input features of the previous one, the derivatives of the errors with respect to the weight are computed as:
\begin{equation}{\label{eq:14}}
    \frac{\partial E}{\partial \omega^{(p)}_{ij}} = \frac{\partial E}{\partial z^{(p)}_{ij}}\frac{\partial z^{(p)}_{i}}{\partial \omega^{(p)}_{ij}} = y^{(p-1)}_{i} \frac{\partial E}{\partial z^{(p)}_{ij}}
\end{equation}

At last, the weights are adjusted as:
\begin{equation}{\label{eq:15}}
    \omega^{(p)}_{ij}|_{t+1} = \omega^{(p)}_{ij}|_{t} - \alpha\frac{\partial E}{\partial\omega^{(p)}_{ij}}
\end{equation}
where $\alpha$ is the learning rate and subscript t denotes the iteration level Commonly available ML softwares provide optimizer which dictate the learning rate. For example, adaptive moment estimation (ADAM) optimizer available in Keras application programing interface (API) requires a user specified initial learning rate, bit the rate adaptively adjusted during training. For the hidden layers a backpropagation is used, where the derivative of information from the hidden layer ahead starting from the output layer, as below:
\begin{equation}
    \frac{\partial E}{\partial y^{(p-1)}_{i}} = \sum_{i=1}^{\infty}\omega^{(p)}_{ij}\frac{\partial E}{\partial z^{(p)}_{i}}
\end{equation}
and then the equation (\ref{eq:14}) and (\ref{eq:15}) are used to obtain $\frac{\partial E}{\partial \omega^{(p-1)}_{ij}}$.

\subsection{\label{sec:level4}Turbulent flows \& Datasets Used}

The neural network was trained validated and tested on 2 different flows for which high fidelity data( DNS and experimental data) were available and the RANS results were obtained in ANSYS Fluent with $k-\epsilon$ modeling and fully developed turbulent flow.

\begin{figure}[hbtp]
    \includegraphics[scale=1]{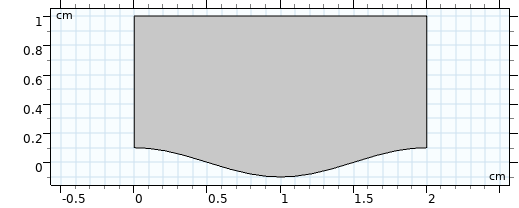}
    \caption{outline of flow domain of wavy wall}
    \label{fig:wavywall(2)}
\end{figure}

The first flow in database was flow over a wavy wall at $Re=6180$. A schematic of the flow domain is shown in Figure \ref{fig:wavywall(2)}. We chose $128*128$ grid over $x$ and $y$ direction respectively, so the total number of grid points were $16384$. 
\begin{figure}[hbtp]
    \hspace{18cm}
    \includegraphics[scale=1]{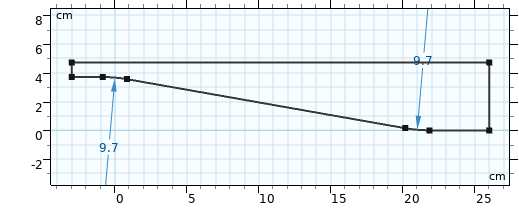}
    \caption{outline of flow domain of Buice 2D diffuser}
    \label{fig:buice diffuser(1)}
\end{figure}




The second case we have considered is Buice diffuser that has been experimentally investigated by \cite{buice2000experimental}. \cite{obi1993experimental} have done an experiment over this diffuser and data from this experiment is available. Fig \ref{fig:buice diffuser(1)} shows the set up they used for their experiment. we are using setup for our simulation so we can use their data as a high fidelity data in our work. All data used in this study were from the National Aeronautics and Space Administration (NASA) Langley Research Center Turbulence Modeling Resource and are available online.

We used the 80\% of wavy wall data for our training phase, 20\% for validation and and we used the Buice 2D diffuser data for testing the final results of our model, in order to see how our model works for flows which it has not seen before. 

The neural network model learns a mapping from the space of local flow features to the target of the magnitude of eigenvalue perturbation. In order for the results to exhibit good generalization properties, the inputs should be rotationally invariant and non-dimensional. In order to construct these inputs, the following raw local flow
variables were considered: the mean pressure $P$ and its gradient vector, the mean velocity $U$ and its gradient tensor, the turbulent kinetic energy $k$ and its gradient vector, the turbulent dissipation rate $\epsilon$, the eddy viscosity $\nu_t$ , the mean density $\rho$, the molecular viscosity $\nu$, and the distance to the nearest wall $d$. Table \ref{tab:1} shows these non-dimensional features.

 \begin{table}[h]
     \resizebox{12 cm}{4 cm}{%
     \hskip-1.8cm
     \begin{tabular}{c|c|c|c|c|c|}
        \# & Description & Formula & \# & Description & Formula\\ \hline\hline 
         1 & \shortstack{\\Non-dimensionalize\\ Q criterion} & $\frac{||R||^{2}- ||S||^{2}}{||R||^{2} + ||S||^{2}}$ & 5 & \shortstack{\\Ratio of pressure \\normal stresses to\\ normal shear stresses} & $\frac{\sqrt{\frac{\partial P}{\partial x_{i}}\frac{\partial P}{\partial x_{i}}}}{\sqrt{\frac{\partial P}{\partial x_{j}}\frac{\partial P}{\partial x_{j}}} + 0.5\rho\frac{\partial U^{2}_{k}}{\partial x_{k}}}$\\ \hline
         2 & Turbulence intensity & $\frac{k}{0.5U_{i}U_{i} + k}$ &  6 & \shortstack{\\Viscosity ratio} & $\frac{\nu_t}{100\nu + \nu_t}$\\ \hline
         3 & \shortstack{\\Turbulence Reynolds
\\number} & $min(\frac{\sqrt{k}d}{50\nu},2)$ & 7 & \shortstack{\\Ratio of total Reynolds\\stresses to normal\\Reynolds stresses} & $\frac{||u^{'}_{i}u^{'}_{j}||}{k + ||u^{'}_{i}u^{'}_{j}||}$ \\ \hline
         4 & \shortstack{\\Pressure gradient\\along streamline} & $\frac{U_{k}\frac{\partial P}{\partial x_{k}}}{\sqrt{\frac{\partial P}{\partial x_{j}}\frac{\partial P}{\partial x_{j}}U_{i}U_{i}} + |U_{l}\frac{\partial P}{\partial x_{l}}|}$ & 8 & \shortstack{\\Mach number} & $\frac{||\overline{U}||_{2}}{C_{0}}$ \\ \hline
         \multicolumn{2}{c}{9} & \multicolumn{2}{c}{\shortstack{\\Ratio of turbulent\\time scale to mean\\strain time scale}} &\multicolumn{2}{c}{ $\frac{||S||k}{||S||k + \epsilon}$}\\ 
         
        \hline\hline
         
     \end{tabular}}
     \caption{Non-dimensional inputs.}
     \label{tab:1}
 \end{table}

In this table, $R_{ij} = \frac{1}{2}(\frac{\partial U_{i}}{\partial x_{j}} - \frac{\partial U_{j}}{\partial x_{i}})$ is the anti-symmetric rotation rate tensor, and $||R||$ is its Frobenius norm. Similarly, $S_{ij} = \frac{1}{2}(\frac{\partial U_{i}}{\partial x_{j}} + \frac{\partial U_{j}}{\partial x_{i}})$ is the symmetric strain rate tensor with Frobenius norm $||S||$. The Reynolds stresses in inputs 7  was calculated using Boussinesq model.

 Our neural network consist of 9 inputs, 8 hidden layers with 15 neurons each layer, and the output is 1 neuron which is the prediction for the discrepancies in the Reynolds stress tensor eigenvalues between DNS and RANS results. Fig \ref{fig:Neural Network} shows our neural network architecture.  The neural network model was trained using the PyTorch library \citep{paszke2017automatic, NEURIPS2019_9015}. We used an Adam optimizer with mini-batching for optimization. The learning rate we used for our model is $\alpha = 2.5 * 10^{-4}$ and we randomly initialized weights and biases for our model based on the Xavier initialization approach. Also, to avoid overfitting in our model we used dropout method in each hidden layer.

\section{\label{sec:level3}Results and Discussion}
  
A neural network was used to predict the amount of eigenvalue perturbation for flow over wavy wall and in a Buice 2D diffuser. We use the neural network to define a functional mapping between the input features and the eigenvalue perturbation. The magnitude of the eigenvalue perturbation was quantified by the magnitude of the Eulerian distance between the RANS simulations and the DNS data on the Barycentric triangle. We compare the results obtained from our model with DNS and experimental data.

\begin{figure}[htbp]
    \centering
    \includegraphics[scale = 1.1 ]{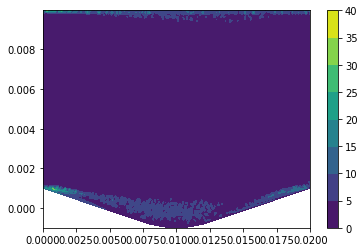}
    \caption{predicted values of eigenvalues perturbation magnitude by neural network for flow over a wavy wall}
    \label{fig:predicted values for wavy wall }
\end{figure}

Fig \ref{fig:predicted values for wavy wall } shows the predicted value of our $\Delta_B$ over the flow domain of the wavy wall case. As we can see, at the bottom of the flow domain our model predicted a large difference between the eigenvalues of Reynolds stress tensor of DNS data and the eigenvalues of Reynolds stress tensor for $k-\epsilon$ model. \cite{segunda2018experimental} have carried out PIV measurements for this experiment and report that this is the region where flow separation takes place. \cite{zilker1979influence} have also reported this in their experiments. The $k-\epsilon$ model can not predict turbulent flow separation accurately \citep{patel1991turbulent}. The predictions of the neural network model agree with physics based knowledge of turbulent flows. 

\begin{figure}[htbp]
    \centering
    \includegraphics[scale = 1.1]{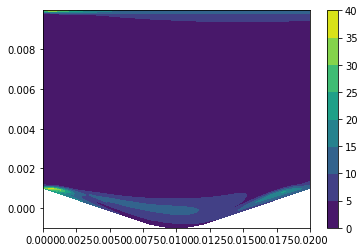}
    \caption{magnitude of eigenvalue discrepancies between RANS prediction and DNS simulations for the wavy wall case}
    \label{fig:neural network True values for wavy wall}
\end{figure}

Fig \ref{fig:neural network True values for wavy wall} shows the True values for eigenvalues perturbation which was obtained by a DNS data. We also see a great difference at place where separation occurs over wavy wall. Comparing the results of our model with the real values we see a good agreement between them. This shows that our model can predict values of perturbation for region of separation. 

\begin{figure}[htbp]
    \centering
    \includegraphics[scale= 1]{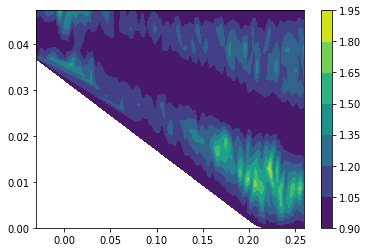}
    \caption{predicted values of eigenvalues perturbation magnitude with neural network in Buice 2D diffuser}
    \label{fig:predicted values for Buice 2D diffuser}
\end{figure}

For our next flow, we have flow in a Buice 2D diffuser. We know that $k-\epsilon$ simulation is unable to predict accurate results for flow in this diffuser because of turbulent flow separation on the bottom wall where divergence occurs. Fig \ref{fig:predicted values for Buice 2D diffuser} shows the True values of perturbation gained by difference between DNS data and RANS simulation. As we observe there are some regions which DNS and RANS simulation are so different from one another, and those regions are where separation occurs in this flow at the ottom wall.

\begin{figure}
    \centering
    \includegraphics{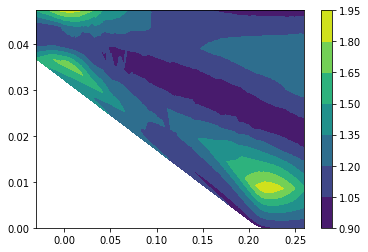}
    \caption{magnitude of eigenvalue discrepancies between RANS prediction and DNS simulations for the Buice 2D diffuser case.}
    \label{fig:RANS sim for diffuser}
\end{figure}

We can compare this results with RANS simulation. Fig \ref{fig:RANS sim for diffuser} shows the predicted value of our neural network for flow over Buice 2D diffuser. Comparison of these two figures shows the accuracy of our model for regions with separation. There are 2 reasons for this the nature of the perturbations used in this paper and the generalizability behavior. Flow separation is best accounted for using perturbations to the eigenvectors for the Reynolds stress tensor. In this paper we are focusing on the perturbations to the eigenvalues of the Reynolds stress tensor and this limits the accuracy. The limited generalizability of the model is due to the complex architecture of the neural network model that leads to a lot of expressive power but also a lot of parameters to be tuned \citep{zhang2018three}. The amount of data used to train this model was not enough to ensure generalizability and needs to be increased and regularization needs to be used in the model training \citep{wu2009new, larsen1994generalization, ying2019overview}. 

\section{\label{sec:level4}Summary \& Future Work}
In this study we proposed a densely connected neural network model to determine the amount of perturbation needed for RANS uncertainty quantification with eigenvalue perturbation. We used a set of 9 features as input of our model and the distances between the DNS and RANS model predictions of the Reynolds stress tensor in barycentric map as our output target. The model predicts the magnitude of eigenvalue perturbation required to the RANS prediction of the Reynolds stress tensor. There were two turbulent flows in our study: flow over a wavy wall and flow through a Buice 2D diffuser.

We saw that our model was capable of predicting better results compared to RANS $k-\epsilon$ model used in the ANSYS Fluent CFD simulations. Then we compare our prediction with direct numerical simulation, large eddy simulation and experimental data and we saw a good agreement between them.

Lastly, different sets of features or different hyperparameters could be made to get a more accurate results from our model. This needs a assessment of the importance of the different input features for the model and to establish the physics captured by the neural network. A study of the optimal hyperparameters using Bayesian Optimization will improve the accuracy of the model predictions. A new model could be suggested for our training phase. The training phase can be made with a bigger dataset than just two, for example, flow over a backward facing step or flow over a periodic hills. Increasing the diversity of the training dataset will lead to better performance and generalizability for the deep learning model. There can be a new study for perturbation on eigenvectors and the turbulent kinetic energy alongside the eigenvalues for future work. We are carrying these investigations into modeling the eigenvectors and the turbulent kinetic energies and will report the results of those investigations. Also we are using explainability tools to find the importance of different input features, correlate this to the physics captured by the deep learning model and to determine new input features that are informative about the missing physics.

\bibliography{sorsamp}

\end{document}